%% file: main.tex
\documentclass[10pt,twocolumn,letterpaper]{article}

\usepackage{wacv}              
\usepackage[accsupp]{axessibility}

\definecolor{wacvblue}{rgb}{0.21,0.49,0.74}
\usepackage[pagebackref,breaklinks,colorlinks,allcolors=wacvblue]{hyperref}

\def\wacvPaperID{528} 
\def\confName{WACV}
\def\confYear{2026}
\newcommand{\OURNAME}{AFRAgent\hskip0.25em}

\title{AFRAgent : An Adaptive Feature Renormalization Based High Resolution
Aware GUI agent}

\author{
Neeraj Anand$^{* \dagger}$ \quad
Rishabh Jain$^{*}$ \quad
Sohan Patnaik$^{*}$ \quad
Balaji Krishnamurthy \quad
Mausoom Sarkar \\
\\
Media and Data Science Research, Adobe \\
}

\date{}

\begin{document}
\maketitle

\begingroup
\renewcommand\thefootnote{}\footnotetext{
\noindent\begin{tabular}{@{}l@{}}
$^{*}$ equal contribution \\
$^{\dagger}$ Contact: {\scriptsize\texttt{{neeraja,rishabhj,soha}@adobe.com}} \\
$^{1}$ Project Webpage: {\href{https://mdsrlab.github.io/2025/11/30/AFRAgent-WACV.html}{Website}}
\end{tabular}
}
\addtocounter{footnote}{-1}
\endgroup

\input{sec/0_abstract}    
\input{sec/1_intro}
\input{sec/2_related}

\input{sec/3_methodology}

\input{sec/4_experiment}

\input{sec/5_result}
\input{sec/6_analysis}
\input{sec/7_conclusion}
\bibliographystyle{ieeenat_fullname}
\bibliography{main}

\end{document}

%% file: sec/0_abstract.tex
\begin{abstract}
There is a growing demand for mobile user interface (UI) automation, driven by its broad applications across industries. With the advent of visual language models (VLMs), GUI automation has progressed from generating text-based instructions for humans to autonomously executing tasks, thus optimizing automation workflows. Recent approaches leverage VLMs for this problem due to their ability to 1) process on-screen content directly, 2) remain independent of device-specific APIs by utilizing human actions (e.g., clicks, typing), and 3) apply real-world contextual knowledge for task understanding. However, these models often have trouble accurately identifying widgets and determining actions due to limited spatial information in vision encoder features. Additionally, top-performing models are often large, requiring extensive training and resulting in inference delays. In this work, we introduce \OURNAME, an instruct-BLIP-based multimodal architecture that achieves superior performance in GUI automation while being less than \textbf{one-fourth} the size of its nearest competitor. To enhance image embeddings in the large language model (LLM) pipeline, we propose an adaptive feature renormalization-based (a token-level affine transformation) technique that effectively enriches low-resolution image embeddings and fuses high-resolution details. We evaluate \OURNAME on Meta-GUI and AITW benchmarks, establishing a new state-of-the-art baseline for smartphone automation.\footnotemark[1] The code is available at {\scriptsize\texttt{https://github.com/neerajanand321/AFRAgent}}
\end{abstract}

%% file: sec/1_intro.tex
\section{Introduction}

\input{plot_tex/flop_accuracy}
Advancements in deep learning have made previously challenging tasks achievable, particularly with the emergence of transformer-based architectures. Progress in natural language processing (NLP), computer vision, and multimodal models has significantly enhanced our capacity to address complex tasks. Among these tasks, GUI automation is crucial due to its wide-ranging real-world applications and substantial economic impact. A model must accomplish a user-defined objective, such as scheduling calendar events or preparing presentations on specific topics on a desktop or mobile operating system. While this problem has been explored even before the emergence of large language models (LLMs) \cite{Wooldridge_Jennings_1995}, the development of LLMs has led to notable improvements in action prediction accuracy \cite{zheng2023synapse}, strategic planning for sequential steps \cite{lee2024explore, wen2023empowering}, and generalization across various environments and applications \cite{wen2023empowering}. Previous approaches, such as verbalization of screen content combined with prompt engineering and fine-tuning \cite{deng2024mind2web,sun-etal-2022-meta}, have leveraged LLMs to predict necessary subsequent actions. However, since user interfaces are fundamentally intended for visual engagement, text-only methods prove inadequate compared to multimodal models incorporating spatial information.

Large visual language models (VLMs) have demonstrated effectiveness across diverse tasks, including optical character recognition (OCR) \cite{bai2023qwen}, image captioning \cite{agrawal2019nocaps}, visual grounding \cite{yu2016modeling, peng2023kosmos}, question answering \cite{antol2015vqa}, and reasoning \cite{chen-etal-2024-measuring}. Recently, VLMs have also demonstrated promising potential in automating user-defined workflows. For instance, Auto-GUI \cite{zhan2023you} introduced a chain-of-thought approach that combines past action steps with future planning to determine the following action. SeeClick \cite{cheng2024seeclick} emphasized the importance of identifying icons and widgets on the screen, pretraining their network with GUI-grounded data to enhance alignment for automation tasks. CogAgent \cite{hong2024cogagent} incorporated both low and high-resolution vision encoders to capture fine-grained details from mobile screenshots, while CoCo-Agent \cite{ma-etal-2024-coco} provided layout information as input to the LLM and simplified action prediction into a more semantic structure to improve performance.

Despite these advances, current methods have several limitations. First, many approaches rely on external tools like OCR and icon detectors \cite{zhang2021screen, sunkara-etal-2022-towards} to transform the interface into text-based inputs for language models. This dependence can introduce inefficiencies, as lengthy inputs increase processing time, risk exceeding model input limits, and may suffer from information loss or conversion errors. Additionally, extracting layout information from third-party applications can be challenging. In contrast, \OURNAME efficiently processes mobile screenshots directly without needing external tools. Secondly, current approaches rely on computationally intensive models with parameter counts ranging from 7B \cite{ma-etal-2024-coco} to 18B \cite{hong2024cogagent}, making edge deployment difficult despite its critical importance for preserving privacy in UI interactions. Our approach significantly reduces this computational burden by utilizing a more efficient 4B parameter model while maintaining competitive performance, making edge deployment more feasible. Thirdly, existing UI automation methods typically use either Llava-style projections \cite{ma-etal-2024-coco, liu2023llava} or learnable query embeddings \cite{liu2024visual} to extract screen information. However, recent research \cite{hu2024bliva, zhang2024beyond} suggests that a combination of direct projection and learnable query embeddings provides enhanced results. Lastly, existing VLMs often incorporate high-resolution information through methods that substantially increase memory and computational overhead, such as increasing the size of the vision encoder \cite{chen2024far}, splitting images into multiple crops \cite{liu2024improved} resulting in additional LLM tokens, or using cross-attention to integrate information at each layer of the LLM \cite{hong2024cogagent}. To address these issues, we propose an Adaptive Feature Renormalization (AFR) technique that effectively merges image embeddings with learned query projections without adding a significant computational burden. Additionally, we enhance the InstructBLIP architecture by integrating AFR to combine low-resolution and high-resolution features via Q-Former, demonstrating the adaptability of our approach. In summary, our work makes the following technical contributions 
\begin{itemize}
    \item We propose Adaptive Feature Renormalization, a novel technique to enrich the learned query projections with the image embeddings effectively.
    \item We adapt InstructBlip architecture to fuse high resolution information into the low-resolution embeddings via AFR.
    \item Our 4-billion parameter model achieves state-of-the-art results on key GUI benchmarks, significantly surpassing larger models and demonstrating superior performance(Fig \ref{fig:flop_acc}).
    \item We provide comprehensive quantitative and qualitative evidence, along with ablation studies, to demonstrate the effectiveness of each component.
\end{itemize}


%% file: plot_tex/flop_accuracy.tex
\begin{figure}[t]
  \begin{minipage}[b]{0.45\textwidth}  
    \centering
    \includegraphics[width=\textwidth]{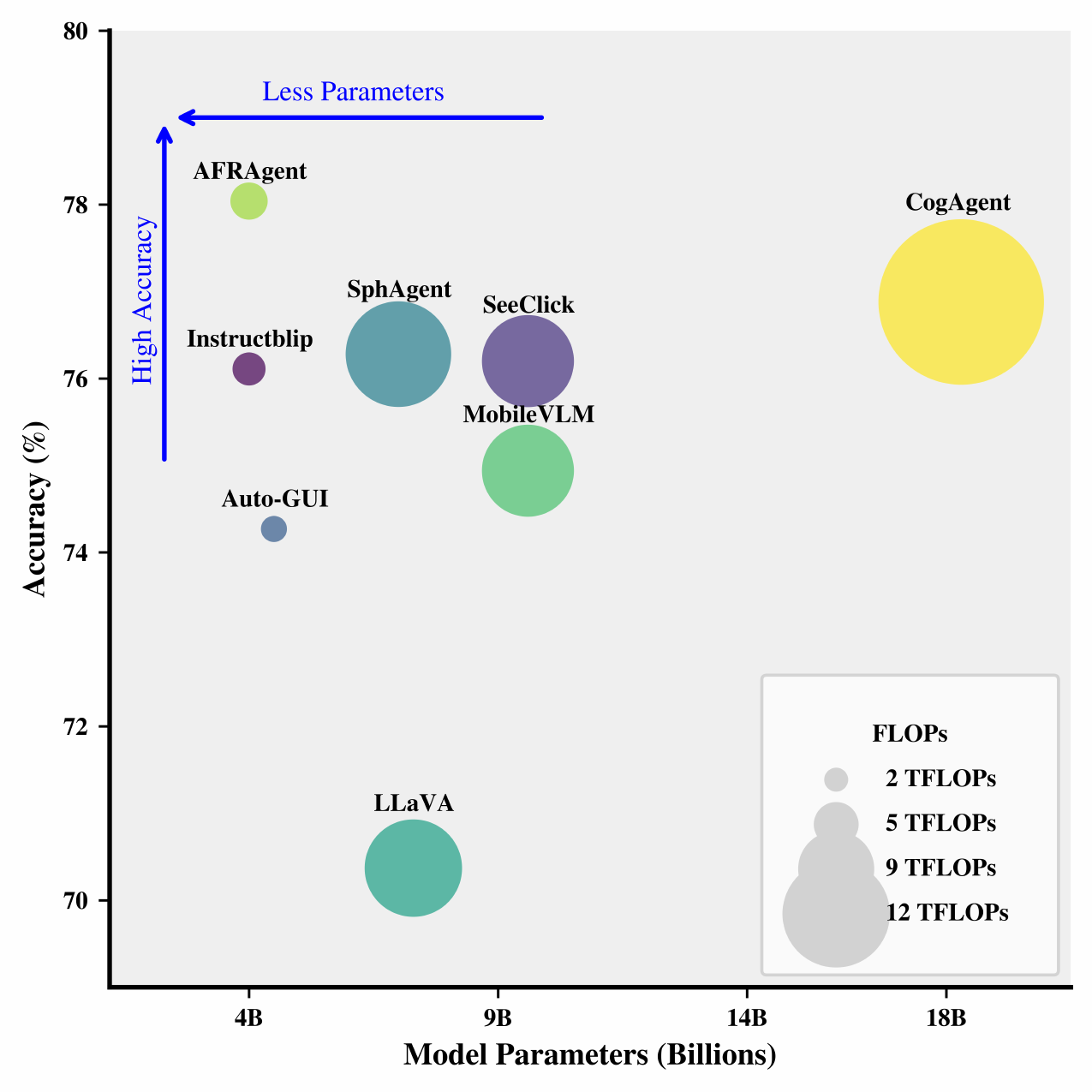}  
  \end{minipage}%
  \hspace{0.04\textwidth}  
  \begin{minipage}[b]{0.48\textwidth}  
    \caption{\OURNAME outperforms existing methods in UI automation on AITW dataset, achieving state-of-the-art results with significantly fewer parameters and reduced computational overhead.}
    \label{fig:flop_acc}
  \end{minipage}
\end{figure}

%% file: sec/2_related.tex
\section{Related Works}

\textit{Agents} are systems that perceive their environment, make decisions, and pursue goals \cite{Wooldridge_Jennings_1995}. Recent advances in language models have improved agents’ adaptability and generalization \cite{osoba2020policy, WANG2020125312}, enabling memory, planning, and applications in domains like software development \cite{li2023camel}, scientific research \cite{boiko2023emergent}, and web-based tasks, especially autonomous GUI navigation using (M)LLMs.

\textbf{Autonomous GUI Navigation} Autonomous navigation in graphical user interfaces (GUIs) leverages these advancements to enable agents to interact with and control complex digital environments. LLM-based methods such as WebAgents \cite{gur2023real}, Mind2Web \cite{deng2024mind2web}, WebArena \cite{zhou2023webarena}, m-BASH \cite{sun-etal-2022-meta}, and AutoWebGLM \cite{10.1145/3637528.3671620} utilize textual representations of screens (e.g., HTML, accessibility trees, OCR) along with task goals to guide navigation. These models effectively process and summarize verbose structures, but they struggle with challenges like HTML verbosity, high token counts, and loss of fine-grained visual details, limiting their ability to fully replicate human-like interactions with GUIs. In contrast, multimodal approaches combine textual and visual information to improve agent performance. Models like WebGum \cite{furuta2023multimodal}, AppAgent \cite{zhang2023appagent}, Mobile-Agent \cite{wang2024mobile}, Auto-GUI \cite{zhan2023you}, UI-TARS\cite{qin2025ui}, and MobileVLM \cite{wu2024mobilevlm} rely solely on screenshots, leveraging visual data to navigate and interact with GUIs naturally. OmniParser \cite{lu2024omniparser} uses GPT-4V and introduces a robust screen parsing technique, achieving state-of-the-art zero-shot performance. CogAgent \cite{hong2024cogagent} incorporates a specialized high-resolution visual module with alignment pre-training, while CoCo-Agent \cite{ma-etal-2024-coco} combines both textual information and screenshots. These multimodal agents overcome many limitations of text-based methods by providing richer context and a deeper understanding of UI elements, though they face challenges in efficiency due to large model sizes and limited high-resolution visual capabilities, reducing effectiveness in tasks requiring precise visual understanding.

\begin{figure*}[t] 
    \centering 
    \includegraphics[width=\textwidth]{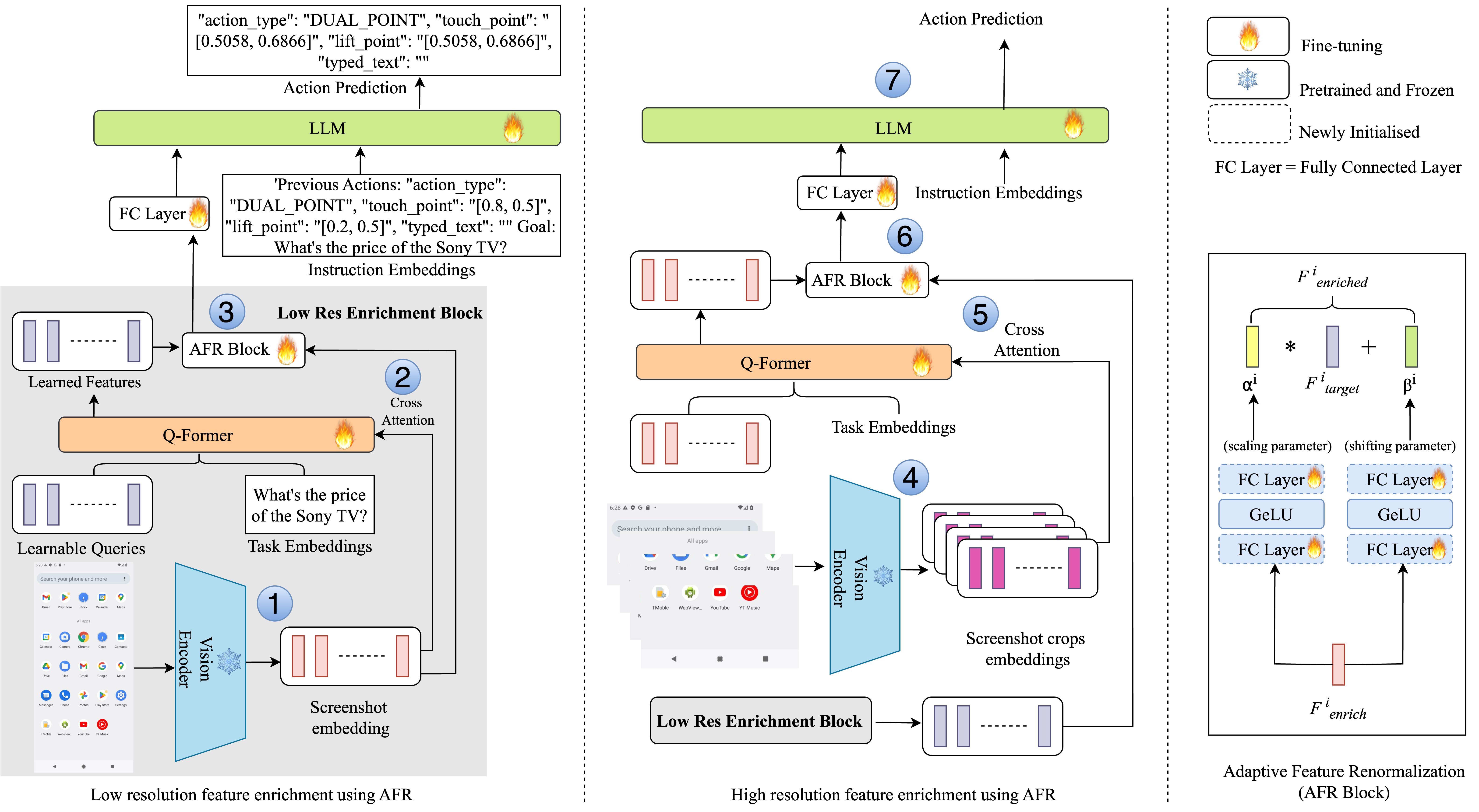} 
     \caption{Core architectural improvements in \OURNAME on the AITW dataset. \textbf{Low-resolution enrichment with AFR:} Image embeddings are generated via BLIP (1) and cross-attended with the Q-Former (2). These control scaling and shifting parameters that renormalize Q-Former features (3). \textbf{High-resolution enrichment with AFR:} The image is divided into crops, processed through BLIP (4) to obtain tokens, which cross-attend with the Q-Former (5) and enrich the low-res features (6). Final enriched features are projected to the LLM for action prediction (7).}
    \label{arch_diagram} 
\end{figure*}

\label{subssec:featureEnchriment}
\textbf{Low-Resolution Feature Enrichment Strategies} Various strategies have been developed to pass information from vision encoders to LLMs, with the most widely used being LLaVA-style projections and the Q-Former-based InstructBLIP architecture. Zhang et al. \cite{zhang2024beyond} demonstrated that while InstructBLIP performs well on tasks like ScienceQA \cite{lu2022learn}, which involves multiple-choice question answering across various science topics, it underperforms on tasks that require deeper reasoning, such as POPE \cite{li2023evaluating}, MME \cite{yin2023survey}, and GQA \cite{hudson2019gqa}, where LLaVA-style direct projection has proven more effective. To balance these strengths, Zhang et al. \cite{zhang2024beyond} proposed a Soft Mixture of Experts (MoE) model with a gating network that learns optimal weights to combine Q-Former and LLaVA-style features. Despite incorporating learnable noise to prevent over-reliance on either feature type, this approach faced challenges with training instability and yielded limited improvements in both performance and generalizability. We also compare our method with the MoE approach in Sec \ref{sec:analysis}.Alternatively, BLIVA \cite{hu2024bliva} uses a projection layer to input both Q-Former embeddings and direct projections into the LLM, but this significantly increases token count and computational costs. Inspired by previous advances in computer vision \cite{huang2017arbitrary, karras2019style, park2019semantic, albahar2021pose}, we introduce Adaptive Feature Renormalization (AFR), which enhances Q-Former features with additional learned embeddings. AFR retains the computational efficiency of the Q-Former approach while capturing the benefits of direct projection, balancing performance with resource efficiency

\textbf{High-Resolution Feature Enrichment Strategies} To merge high-resolution information, existing methods \cite{liu2024improved, chen2024far, shi2025we} often divide the image into smaller regions, process each independently through the vision encoder, and then pass all resulting embeddings to the LLM via a projection layer. Although effective, this approach can become computationally expensive with large LLMs due to the increase in token count. For example, InternVL \cite{chen2024far} introduces up to 3k visual tokens during training and 10k during inference, adding substantial computational burden. To mitigate these costs, SliME \cite{zhang2024beyond} uses a local compression layer with a Q-Former and a text-guided patch-selection router, while CogAgent \cite{hong2024cogagent} employs a high-resolution cross module that injects features via cross-attention layers. AFR, in contrast, provides a simpler, compute-efficient solution by enriching existing features with high-resolution details without excessive overhead.

%% file: sec/3_methodology.tex
\section{Methodology}
\label{sec:method}

\paragraph{UI Automation Task:} GUI automation task requires the agent to execute a series of actions to achieve a specific objective. Formally, consider the GUI automation task \( T \), defined as a free-form text instruction provided by the user. The current state of the user interface is captured as a screenshot, denoted by \( S_t \), where \( t \) represents the time step in the sequence of actions. The agent interacts with the GUI environment through a series of actions \( A = \{a_1, a_2, \dots, a_t\} \), where each action \( a_t \in \mathcal{A}_{\text{UI}}\) (where $\mathcal{A_{\text{UI}}} = \{$click, type, scroll, swipe, press back, press home, press enter, task complete$\}$) ~\citep{zhan2023you} is taken at time step \( t \), transitioning the system from state \( S_{t-1} \) to state \( S_t \). Given the task \( T \), the current state \( S_t \), and the history of actions taken up to time step \( t \), denoted as \( H_t = \{a_1, a_2, \dots, a_t\} \), the goal of the model is to predict the following action \( a_{t+1} \). This can be formalized as learning a policy $\pi(a_{t+1} | T, S_t, H_t)$ which outputs the probability distribution over possible action space $\mathcal{A_{\text{UI}}}$ given the task, current state, and action history. The process continues iteratively until the agent reaches a terminal state \( S_g \), where \( S_g \) satisfies the conditions defined by the task \( T \).

\paragraph{Architecture of AFRAgent:}
\label{sec:arch}
We leverage InstructBLIP backbone for the task of GUI Automation. Consider the image encoder $E$, the QueryFormer $Q$, and the large language model $L$ as the core components of the architecture of \OURNAME\ $M_{\text{Agent}}$. The image encoder \( E \) takes the current screenshot \( S_t \) as input and outputs a set of visual embeddings \( I_t \) by dividing the input image into patches. Formally,
\begin{equation}
  I_t = \left[i_1, i_2, \dots, i_N\right] = E(S_t),
\end{equation}
where $i_n \in \mathbb{R}^{d_I}$, $n = 1, 2, \dots, N$ and \( N \) is the number of visual patches of $S_t$. The Q-Former $Q$ generates instruction-aware visual representations by attending to both visual embeddings and task instructions. Let \( \mathcal{T}_{Q} = [e_{q_1}^{(0)}, e_{q_2}^{(0)}, \dots, e_{q_M}^{(0)}] \) represent the learnable query token embeddings, where each query token \( t_{q_m} \in \mathbb{R}^{d_Q} \) has dimension \( d_Q \). Additionally, we embed the goal task instruction $T$ and the action history $H_t$ with a learnable embedding layer \( g_Q(T, H_t) = [e_1^{(0)}, e_2^{(0)}, \dots, e_{L_T}^{(0)}] \), where each token embedding \( e_l^{(0)} \in \mathbb{R}^{d_Q} \), and \( L_T \) is the number of tokens in the verbalized goal task instruction and the action history. We concatenate the query token embeddings \( \mathcal{T}_{Q} \) with the goal task and action history embeddings \( g_Q(T, H_t) \), forming a combined input to the Q-Former denoted by the following equation.
\begin{multline}
X_{Q}^{(0)} = [E_Q^{(0)};E_T^{(0)}] = [e_{q_1}^{(0)}, e_{q_2}^{(0)}, \dots, e_{q_M}^{(0)}; \\
e_1^{(0)}, e_2^{(0)}, \dots, e_{L_T}^{(0)}]
\end{multline}
where \( X_{Q} \in \mathbb{R}^{(M + L_T) \times d_Q} \). The Q-Former then takes this concatenated input \( X_{Q} \) along with the visual embeddings \( I_t = [i_1, i_2, \dots, i_N] \), and employs a cross attention layer and attends to the user task embeddings using self-attention layers followed by a feed-forward network (FFN), repeated for $Z$ layers. These operations help extract the relevant image features useful for the particular task from the original image embeddings. The following equations describe the operation of Q-former.

\begin{align}
    E_Q^{(z-1)}, E_T^{(z-1)} &= SelfAttention(E_Q^{(z-1)}, E_T^{(z-1)}), \nonumber \\
    E_{Q_{\text{cross}}}^{(z)} &= CrossAttention(E_Q^{(z-1)}, I_t)), \nonumber \\
    E_Q^{(z)}, E_T^{(z)} &= \mathit{FFN}(E_{Q_{\text{cross}}}^{(z)}), \mathit{FFN}(E_T^{(z-1)}),
    \label{eq:qformer}
\end{align}

where $z = 1, 2, \cdots, Z$ denote the layer index. The query token representations thus obtained at the last layer ($Z$) is aligned to the LLM embedding space through a learnable projection layer. Further, we concatenate the projected Q-Former output from \( E_Q^{(Z)} \) i.e., the output after the projection layer, with the goal task and action history embeddings $g_L(T, H_t)$ of the underlying LLM $L$, forming the final input to the LLM:

\begin{equation}
\mathit{X_{LLM}} = [e_{q_1}^{Z}, e_{q_2}^{Z}, \dots, e_{q_M}^{Z}; e_1^{L}, e_2^{L}, \dots, e_{L_T}^{L}],
\end{equation}

where \( X_{LLM}  \in  \mathbb{R}^{(M + L_T)  \times  d_L}  \). Given this as the input, the LLM $L$ outputs the next predicted action $\hat{a}_{t+1} = L(X_{LLM})$. Given the predicted action and the ground-truth action, we enforce the standard cross-entropy loss to train the agent $M_{\text{Agent}}$(Fig \ref{arch_diagram}).

\paragraph{Adaptive Feature Renormalization:} As discussed in Sec \ref{subssec:featureEnchriment}, combining LLAVA-style features with Q-Former features demonstrates superior performance on various multi-modal reasoning benchmarks. Inspired by this observation, we introduce an Adaptive Feature Renormalization (AFR) technique to enhance the input representation for the LLM. AFR leverages both image embeddings and Q-Former outputs to enrich the final LLM input representation of the image.

The AFR block takes as input two sets of features: 1) the enriching features, denoted as \( F_{\text{enrich}} \in \mathbb{R}^{M \times d_Q} \), and 2) the target features to be enriched, denoted as \( F_{\text{target}} \in \mathbb{R}^{M \times d_Q} \). Formally, AFR computes two learnable parameters, \( \alpha \) and \( \beta \), where \( \alpha, \beta \in \mathbb{R}^{M \times d_Q} \), that scale and shift \( F_{\text{target}} \) based on \( F_{\text{enrich}} \), producing the enriched features \( F_{\text{enriched}}  \in  \mathbb{R}^{M  \times  d_Q}  \). This operation is depicted by the following equation:
\begin{align}
     \alpha,  \beta &=  \mathit{FFN_{\alpha}}(F_{\text{enrich}}), \mathit{FFN_{\beta}}(F_{\text{enrich}}),  \nonumber  \\
     F_{\text{enriched}} &= (\alpha  \odot  F_{\text{target}})  \oplus  \beta,
\end{align}
\paragraph{Motivation behind AFR:} Affine transformations have been widely utilized in prior work for feature modulation. For instance, StyleGAN~\citep{huang2017arbitrary, karras2019style} employs Adaptive Instance Normalization (AdaIN) to transfer style content in GAN-based image generation, while SPADE~\citep{park2019semantic} applies Spatial Adaptive Normalization for semantic image synthesis.  Despite their diverse applications, these methods highlight the versatility of affine transformations in injecting information through feature modulation. By scaling and shifting the target features based on the enriching features, we hypothesize that the information from the global enriching features is effectively injected into each target token, effectively renormalizing them and leading to a more informative representation for downstream tasks. We also verify this intuition using Grad-Cam \cite{selvaraju2017grad} visualizations in Sec \ref{sec:analysis}. Next, we describe the integration of Low-Res Image Embeddings and High-Res representations using AFR to obtain enriched features for UI automation.

\paragraph{Low-resolution Enrichment using AFR:} We inject the image embeddings into the Q-Former representations using the AFR operation. As defined in Sec \ref{sec:arch}, the image embeddings \( I_t \) for a given UI screenshot \( S_t \) serve as the enriching features \( F_{\text{enrich}} \), while the Q-Former representations of the query tokens \( E_Q^{(Z)} \) act as the target features \( F_{\text{target}} \). We compute the scaling and shifting parameters \( \alpha^{\text{Image}} \) and \( \beta^{\text{Image}} \in \mathbb{R}^{M \times d_Q} \), where \( M \) is the number of output query token representations and \( d_Q \) is their dimensionality. The enriched Q-Former features, denoted by \( E_Q^{\text{Image}} \in \mathbb{R}^{M \times d_Q} \), are computed as follows:

\begin{align}
    \alpha^{\text{Image}}, \beta^{\text{Image}} &= \mathit{FFN_{\alpha}}(I_t), \mathit{FFN_{\beta}}(I_t)  \nonumber,  \\
    E_Q^{\text{Image}} &= (\alpha^{\text{Image}}  \odot  E_Q^{(Z)})  \oplus  \beta^{\text{Image}},
\end{align}

In our case, patch embeddings or learned queries capture fine-grained semantic details of specific regions on the smartphone screen. By modulating these core features before feeding them into the LLM, we inject valuable information while preserving the original strengths of InstructBLIP's pre-trained features.
\paragraph{High-resolution Enrichment using AFR:} Through empirical evidence, we found that most errors in action recognition for UI automation stem from the limited input resolution of the image encoder. This highlights the need for high-resolution features within the visual language model (VLM) to enhance accuracy. Existing methods ~\citep{liu2024improved, chen2024far, shi2025we} often address this by dividing the image into smaller sections, processing each section independently through the vision encoder, and passing all resulting embeddings to the LLM via a projection layer. This setting can become prohibitively expensive computationally in the case of large LLMs due to the additional token count. Therefore, we leverage the AFR Block in order to fuse the high resolution information onto the Q-Former representation enriched with image embeddings. For a screenshot $S_t$, we obtain the high resolution image $\Tilde{S}_t$ and split it into $C$ crops horizontally $S_t = [S_t^{(1)}, S_t^{(2)}, \cdots, S_t^{(C)}]$ to better preserve the aspect ratio during resizing. We pass each of the crops to the vision encoder to obtain the image embeddings denoted by $\Tilde{I_t}^{(c)} = [\Tilde{i}_1^{(c)}, \Tilde{i}_2^{(c)}\cdots, \Tilde{i}_N^{(c)}]$. Since the query tokens defined earlier were used to process the low resolution image, to capture fine-grained information via the high resolution image embeddings, we introduce a new set of learnable query token embeddings denoted by $\Tilde{\mathcal{T}}_Q = [\Tilde{e}_{q_1}^{(0)},  \Tilde{e}_{q_2}^{(0)}, \cdots, \Tilde{e}_{q_M}^{(0)}]$ whereas the goal task and action history embeddings remain the same $g_Q(T, H_t)$. Using the same Q-Former $Q$, we obtain the high-resolution output query token representation $\Tilde{E}_Q^{Z}$ using the following equation. 
\begin{equation}
    \Tilde{E}_Q^{(Z)}, \Tilde{E}_T^{(Z)} = Q(\Tilde{E}_Q^{(0)}; E_T^{(0)}; [\Tilde{I}_t^{(1)}, \Tilde{I}_t^{(2)} \cdots \Tilde{I}_t^{(C)}]),
\end{equation}
where we feed the new query token embeddings for high resolution, the task and action history embeddings, and concatenate the image embeddings of each of the crops as the input to the Q-Former. Using the AFR Block, we use $\Tilde{E}_Q^{(Z)}$ as $F_{\text{enrich}}$ and $E_Q^{\text{Image}}$ as $F_{\text{target}}$ to obtain $E_Q^{\text{High}}$, denoted by the following equation.
\begin{align}
    \alpha^{\text{High}}, \beta^{\text{High}} &= \mathit{FFN_{\alpha}}(\Tilde{E}_Q^{(Z)}), \mathit{FFN_{\beta}}(\Tilde{E}_Q^{(Z)}),  \nonumber  \\
    E_Q^{\text{High}} &= (\alpha^{\text{High}}  \odot  E_Q^{\text{Image}})  \oplus  \beta^{\text{Image}},
\end{align}
After obtaining the final enriched features $E_Q^{\text{High}}$, we apply the projection layer on top and pass the visual features, task instruction and action history to the LLM for output generation. The model is trained using the language modelling loss on the predicted output as discussed in Sec \ref{sec:arch}.

%% file: sec/4_experiment.tex
\section{Experiments}

This section describes our experimental setup, covering datasets, baselines, and evaluation metrics. Additional details are in the supplementary material. We evaluate our approach on two key datasets: Meta-GUI and AITW, which span diverse mobile GUI interaction tasks.

\paragraph{Datasets} \textbf{Meta-GUI} \cite{sun2022meta} consists of 1k episodes with 18k steps across 11 applications in 6 domains, including weather, calendar, and search. Each episode is structured as a multi-turn dialogue, with agents interacting with users and confirming or discussing next steps. We follow CoCo-Agent \cite{ma-etal-2024-coco} by incorporating dialogue and action history as input. Additional details can be found in supplementary. \textbf{AITW} \cite{rawles2023androidinthewild} is the largest dataset for smartphone GUIs, containing 715k episodes and 30k unique instructions across Android apps and websites. It includes multi-step categories (GoogleApps, Install, WebShopping, General) and a single-step category (Single), with each episode comprising a goal instruction and a sequence of screenshot-action pairs.

\paragraph{Experiment Settings} For Meta-GUI, we use the same settings as Meta-GUI \cite{sun2022meta} and CoCo-Agent \cite{ma-etal-2024-coco}. We follow the standard (80/10/10) training/validation/test split of AITW used by Auto-GUI \cite{zhan2023you}, SphAgent \cite{chai2024amex}, and CogAgent\cite{hong2024cogagent} for fair comparison. Importantly, we finetune our model  \textbf{without any UI pretraining}, unlike CogAgent \cite{hong2024cogagent}, SeeClick \cite{cheng2024seeclick}, and SphAgent \cite{chai2024amex}, demonstrating the robustness of our approach to outperform these methods without prior GUI grounding. 

\input{tables/meta_gui_results}

\paragraph{Baselines and Implementation} We compare \OURNAME with diverse baselines spanning different architectures and training paradigms, including multimodal methods using textual (HTML, accessibility trees) and visual inputs (screenshots). Key baselines include \textbf{MM-Navigator}~\cite{yan2023gpt}, \textbf{LLaMA-2}~\cite{touvron2023llama}, \textbf{Auto-GUI}~\cite{zhan2023you}, \textbf{CogAgent}~\cite{hong2024cogagent}, \textbf{CoCo-Agent}~\cite{ma-etal-2024-coco}, \textbf{MobileVLM}~\cite{wu2024mobilevlm}, and \textbf{SphAgent}~\cite{chai2024amex}. On Meta-GUI, we also compare with \textbf{LayoutLM}~\cite{xu2020layoutlm}, \textbf{BERT}, \textbf{m-BASH}~\cite{sun2022meta}, and \textbf{CoCo-Agent}. LLaMA-2 and BERT are pure language models, while MM-Navigator uses few-shot GPT-4V. The rest are fine-tuned multimodal models. Notably, CogAgent \cite{hong2024cogagent} utilizes high-resolution ViT tokens that directly cross-attend with its LLM, whereas SphAgent \cite{chai2024amex} feeds high-resolution image tokens directly into the LLM. \OURNAME is trained for 12 epochs using Adam (lr=$5\times10^{-5}$), with 257 query tokens ($M$), Q-Former hidden size 768 ($d_Q$), 256 image patches ($N$), and LLM hidden size 2048. Screenshots are divided into 4 sections ($C$), and 8 past actions ($H_t$) are used as history. The AFR block has two fully connected layers with GeLU activation. All experiments were conducted on 8xA100 80GB GPUs. We found that AFR introduces no noticeable instability during training, and the model converges steadily without requiring special initialization. We observed consistent improvements in performance starting from the early epochs, indicating effective feature fusion from the AFR mechanism.

\input{tables/aitw_results}

\paragraph{Evaluation Metrics} We use action matching \cite{rawles2023androidinthewild} as our main metric to assess alignment with ground-truth actions. Click actions are considered matches if they are within a 14\% screen distance or the same bounding box as the target UI element, while scrolls match if they share the same primary scroll axis (vertical or horizontal). Exact match accuracy is used for other action types, except \textit{typed\_text} and \textit{dialogue} response. On AITW, predicted text is correct if it contains the label; on Meta-GUI, we use Action Completion Rate(CR) for accuracy and F1 score for input text matching. An action is correct only if all components (e.g., scroll direction, typed text) are accurately predicted.

%% file: tables/meta_gui_results.tex
\begin{table*}[ht]
    \centering
    \resizebox{0.93\textwidth}{!}{
        \begin{tabular}{lccccccccc}
            \toprule
            \textbf{Method} &
            \textbf{Params} & \textbf{Act. Type} & \textbf{Item Acc.} & \textbf{Direction Acc.} & \textbf{Utter. (BLEU)} & \textbf{Input (F1)}  & \textbf{Input (EM)} & \textbf{Action (CR)}\\
            \midrule
            LayoutLM \cite{xu2020layoutlm} &  343M & 82.22 & 71.98 & 94.87 & 50.43 & 90.56 & 83.04 &  67.76  \\
            LayoutLM$_{\text{v2}}$ &  426M & 85.60 & 64.38 & 92.95 & 58.20 & 70.76 & 47.37 &  64.48 \\
            BERT \cite{devlin2018bert} & 340M & 87.52 & 82.84 & 93.59 & 62.19 & 97.24 & 93.57 &  78.42 \\
            LLaVA \cite{liu2023llava} & 7.3B & 87.47 & 77.49 & 98.18 & 67.24  & 96.06 & - &   76.27 \\
            LLaVA$_{\text{w/history}}$ & 7.3B & 91.68 & 81.23 & 97.62 & 66.57  & 96.93 & -  &  81.08 \\
            m-BASH \cite{sun2022meta} & 340M & 90.80 & 85.90 & 96.42 & 63.11 & 94.23 & 91.23 &  82.74  \\
            CoCo-Agent \cite{ma-etal-2024-coco} & 7.3B & 92.59 & 91.72 & \textbf{98.39} & 65.90 & 96.15 & - &  88.27  \\
            AFRAgent & 4B & \textbf{93.28} & \textbf{95.06} & 97.02 & \textbf{67.6} & \textbf{97.94} & \textbf{94.44}  & \textbf{90.83} \\
            \bottomrule
         \end{tabular}
    }
    \caption{Comparative performance on Meta-GUI benchmark evaluating Completion rate (CR), action type, item/direction accuracy, text metrics (F1, exact match), and response quality (BLEU). \OURNAME achieves state-of-the-art accuracy across most metrics, showing efficiency with fewer parameters and improved input handling.}
    \label{tab:meta_gui_results}
\end{table*}

%% file: tables/aitw_results.tex
\begin{table*}[ht]
    \centering
    \resizebox{0.84\textwidth}{!}{
        \begin{tabular}{lcccccccc}
            \toprule
            \textbf{Method} & 
            \textbf{Params} & \textbf{General} & \textbf{Install} & \textbf{GoogleApps} & \textbf{Single} & \textbf{WebShop.} &\textbf{Overall}\\
            \midrule
            \multicolumn{8}{c}{\textbf{Structured Layout Setting}} \\ 
            \midrule
            CoCo-Agent\cite{ma-etal-2024-coco} & 7.3B &    70.96 & 81.46 & 76.45 & 91.41 & 75.00 & 79.05 \\
            AFRAgent & 4B & 71.62 & 80.81 & 76.26 & 90.78 & 75.10 & 78.92\\
            \midrule
            \multicolumn{8}{c}{\textbf{Pure Multimodal Setting}} \\ 
            \midrule
            MM-Navigator$^*$\cite{yan2023gpt} & -  & 43.01 & 46.14 & 49.18 & 78.29 & 48.18 & 52.96\\
            LLaMA-2 \cite{touvron2023llama} & 7B & 28.56 & 35.18 & 30.99 & 27.35 & 19.92 & 28.40\\
            Auto-GUI \cite{zhan2023you} & 4.5B & 68.24 & 76.89 & 71.37 & 84.58 & 70.26 & 74.27\\
            LLaVA \cite{liu2023llava} & 7.3B  & 58.93 & 72.41 & 70.81 & 83.73 & 65.98 & 70.37 \\
            MobileVLM \cite{wu2024mobilevlm} & 9.6B  & 69.58 & 79.87 & 74.72 & 81.24 & 71.70 &  74.94\\
            InstructBlip \cite{liu2024visual}&  4B & 70.66 & 79.59 & 73.05 & 84.99 & 72.26 &   76.11 \\
            SeeClick \cite{cheng2024seeclick} & 9.6B & 67.6 & 79.6 & \textbf{75.9} & 84.6 & 73.1 &   76.2 \\
            SphAgent \cite{chai2024amex} & 7B  & 68.2 & 80.5 & 73.3 & 85.4 & \textbf{74} & 76.28\\
            ShowUI \cite{lin2025showui} & 2B  & 63.9 & 72.5 & 69.7 & 77.5 & 66.6 & 70\\
            CogAgent \cite{hong2024cogagent}& 18.3B & 65.38 & 78.86 & 74.95 & \textbf{93.49} & 71.73 &  76.88 \\
            AFRAgent & 4B & \textbf{70.67} & \textbf{80.89} & 74.16 & 91.06 & 73.27 & \textbf{78.01}\\
            \bottomrule
        \end{tabular}
    }
    \caption{Comparison of \OURNAME and prior models on AITW with unified training. \OURNAME excels in the multimodal setting and performs competitively in the structured layout setting, achieving strong action accuracy with lower compute and memory. $^*$ indicates few-shot setting.}
    \label{tab:aitw_results}
\end{table*}

%% file: sec/5_result.tex
\section{Results}

\input{plot_tex/click_comparison}

We assess \OURNAME on two benchmark datasets, Meta-GUI and AITW, spanning a broad range of UI automation tasks. On the Meta-GUI dataset (Tab.~\ref{tab:meta_gui_results}), \OURNAME establishes a new state-of-the-art, significantly improving upon existing methods across multiple key metrics. Specifically, we achieve a \textbf{2.56\%} increase in action completion rate and a substantial \textbf{3.34\%} improvement in target item prediction accuracy. Additionally, our model excels in input text matching metrics, underscoring its enhanced task comprehension and efficiency despite having fewer parameters.

AITW, a widely used large-scale Android benchmark, serves as a comprehensive testbed for evaluating our model. We report results under two settings: \textbf{Structured Layout Setting} augments text inputs with layout data from IconNet~\cite{sunkara2022towards}, incurring high computational cost ($\sim$1600 extra tokens/screen), an additional inference pass, and susceptibility to error propagation, limiting generalization across UIs. \textbf{Purely Multimodal Setting} evaluates VLMs using only screenshots and natural language goals, avoiding external tools or layout data. It has emerged as the preferred evaluation method, reflecting a model’s intrinsic visual-semantic understanding. Despite being \textbf{78\% smaller}, \OURNAME achieves superior overall accuracy in the multimodal setting, outperforming prior methods across most tasks. It also outperforms CoCo-Agent~\cite{ma-etal-2024-coco} on general and webshop splits in the structured setting, with significantly lower memory overhead.

Figure~\ref{fig:click_comparison} further illustrates qualitative gains, demonstrating enhanced spatial awareness and task comprehension achieved by \OURNAME's novel AFR block integration. In row (a), \OURNAME accurately identifies the relevant icon due to enhanced spatial perception from our high-resolution embedding integration. In row (b), \OURNAME precisely identifies the correct action and input text, highlighting improved task comprehension from rich spatial embeddings.




%% file: plot_tex/click_comparison.tex
\begin{figure}[t]
  \centering
  \includegraphics[width=0.9\linewidth]{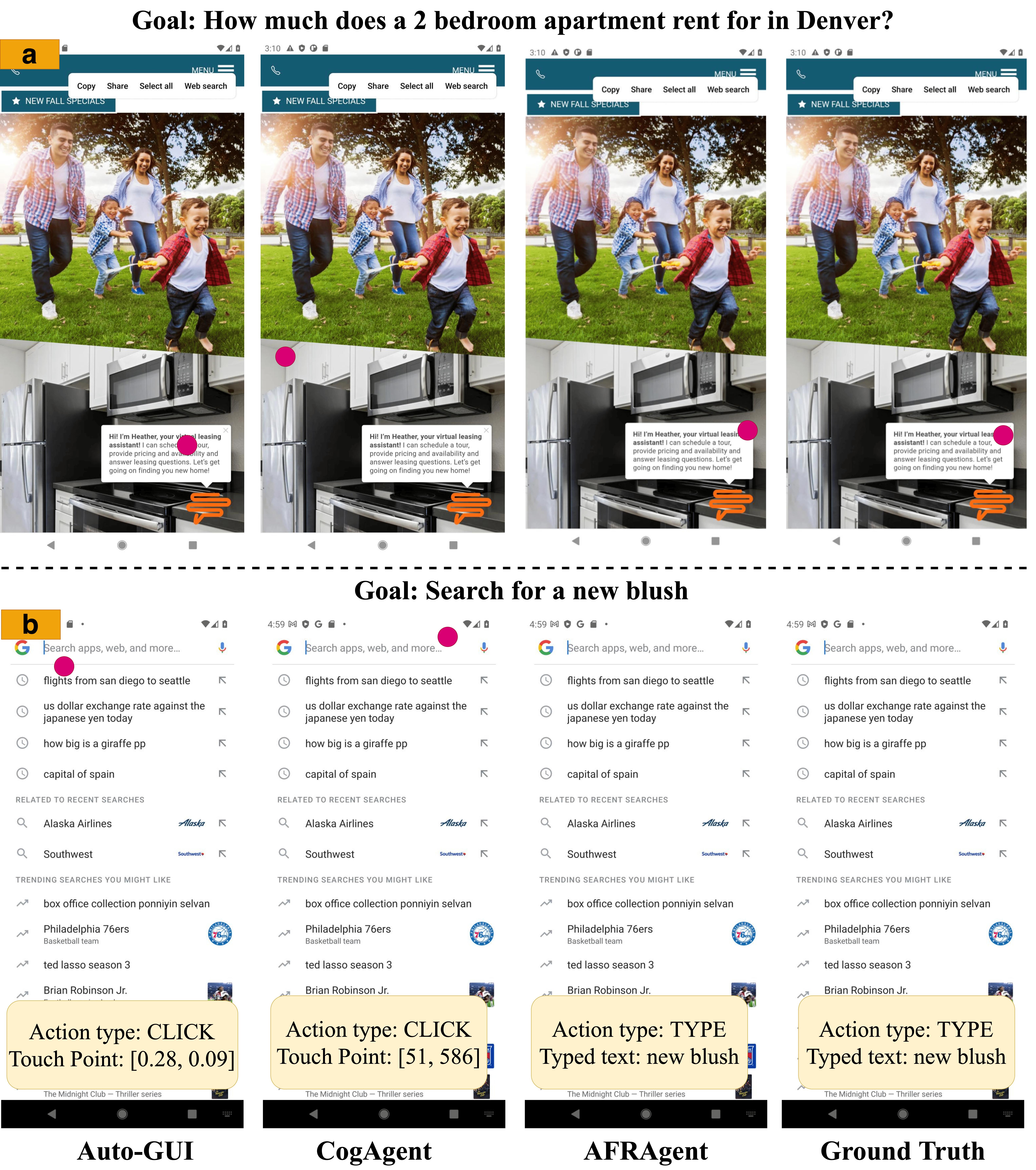}
  \caption{Qualitative comparison of various methods for predicting next action. The figure shows the scenarios where the agent is a) required to click the close icon of a pop-up window and b) type ``new blush" in the search bar. While other methods failed, our method successfully predicted the correct coordinate for clicking and inputted the correct text, highlighting the superior performance of our approach.}
  \label{fig:click_comparison}
\end{figure}

%% file: sec/6_analysis.tex
\section{Analysis}
\label{sec:analysis}

\begin{table}[h!]
    \centering
    \resizebox{\linewidth}{!}{
        \begin{tabular}{lccccccccc}
        \toprule
        \textbf{LVLMs} & \textbf{Params} & 
        \multicolumn{2}{c}{\textbf{Mobile}} & 
        \multicolumn{2}{c}{\textbf{Desktop}} & 
        \multicolumn{2}{c}{\textbf{Web}} & 
        \textbf{Average} \\
        \cmidrule(r){3-4} \cmidrule(r){5-6} \cmidrule(r){7-8}
        & & \textbf{Text} & \textbf{Icon/Widget} & \textbf{Text} & \textbf{Icon/Widget} & \textbf{Text} & \textbf{Icon/Widget} & \\
        \midrule
        MiniGPT-v2*\cite{chen2023minigpt} & 7B & 8.4\% & 6.6\% & 6.2\% & 2.9\% & 6.5\% & 3.4\% & 5.7\% \\
        Qwen-VL*\cite{bai2023qwen} & 9.6B & 9.5\% & 4.8\% & 5.7\% & 5.0\% & 3.5\% & 2.4\% & 5.2\% \\
        GPT-4V* & - & 22.6\% & 24.5\% & 20.2\% & 11.8\% & 9.2\% & 8.8\% & 16.2\% \\
        \midrule
        Fuyu\cite{bavishi2023introducing} & 8B & 41.0\% & 1.3\% & 33.0\% & 3.6\% & 33.9\% & 4.4\% & 19.5\% \\
        CogAgent\cite{hong2024cogagent} & 18.3B & 67.0\% & 24.0\% & \textbf{74.2\%} & 20.0\% & \textbf{70.4\%} & 28.6\% & 47.4\% \\
        SeeClick\cite{cheng2024seeclick} & 9.6B & 78.0\% & 52.0\% & 72.2\% & 30.0\% & 55.7\% & 32.5\% & 53.4\% \\
        AFRAgent & \textbf{4B} & \textbf{78.52\%} & \textbf{53.14\%} & 72.62\% & \textbf{31.44\%} & 64.51\% & \textbf{33.44\%} & \textbf{55.61\%} \\
        \bottomrule
    \end{tabular}
    }
    \caption{Performance comparison across Mobile, Desktop, and Web scenarios for Text and Icon/Widget elements on \textit{ScreenSpot}\cite{cheng2024seeclick}.* represent non GUI-specific methods}
    \label{tab:screenspot_performance}
\end{table}

\paragraph{Generalization to Web/Desktop:} We pre-trained our model on SeeClick \cite{cheng2024seeclick} data before testing the cross-platform generalization on \textit{ScreenSpot} \cite{cheng2024seeclick} dataset. This dataset evaluates VLM's ability to locate text and icons across platforms by measuring click accuracy. We crop screenshots in the original 4$\times$1 grid for mobile training and a 2$\times$2 grid for web/desktop data. AFRAgent performs better than Fuyu \cite{bavishi2023introducing}, CogAgent, SeeClick, etc. across all platforms, as observed in Tab \ref{tab:screenspot_performance}. However, CogAgent performs better in text inputs due to its more extensive pretraining data and greater model size. 

\paragraph{Grad-CAM Analysis}  
To evaluate the impact of feature enrichment, we visualize attention from the 18th layer of the vision encoder using Grad-CAM \cite{selvaraju2017grad} averaged across the entire LLM response and encoder attention heads. As seen in Fig. \ref{fig:gradCam}, AFRAgent effectively renormalizes features, allowing the model to focus on key regions after enrichment. For example, in row (b), the network correctly identifies the target product on Amazon following high-res fusion.

\input{tables/computational_analysis}



\paragraph{Computational Complexity Analysis}
The compute overhead introduced by enriching low- and high-resolution features using AFR is analyzed in Table~\ref{tab:complexity_analysis}. \OURNAME\textsubscript{Lowres} runs nearly 3$\times$ faster with 54\% fewer FLOPs than CogAgent, thanks to its lightweight AFR-based low-resolution enrichment. While $AFR_{HighRes}$ has \(\sim70\%\) cost over $AFR_{LowRes}$, both remain far cheaper than CogAgent, and $AFR_{LowRes}$ already improves over baselines(Additional details in supplementary). High-resolution enrichment increases computation due to additional ViT passes and cross-attention between high-resolution features and the Q-Former. The complexity of this cross-attention operation is:
\[
\mathcal{O}\left((M + L_{\text{T}}) \times C \times N_{\text{N}} \times H_{\text{cross}} \times d_{\text{Q}}\right)
\]
where $M$, $L_{\text{T}}$, $C$, $N_{\text{N}}$, $H_{\text{cross}}$, and $d_{\text{Q}}$ are query or text tokens, image crops, tokens per image, cross-attention heads, and Q-Former embedding size, respectively. Notably, our approach is much more efficient than CogAgent, which applies cross-attention to a large LLM decoder, while we leverage the compact Q-Former.

\input{plot_tex/grad_cam}

\input{tables/Different_fusion}
\paragraph{Comparison of Fusion Strategies}  
We evaluate the effectiveness of AFR as a fusion strategy in low-resolution and high-resolution settings on the General and Single subsets of AITW. For low-resolution fusion, we compare AFR against a residual approach, defined as \( F_{enriched} = F_{target} + \text{MLP}(F_{enrich}) \), and the Mixture of Experts (MoE) method \cite{zhang2024beyond}. For the high-resolution setting, we assess Qwen2-VL using the \textit{AnyRes} configuration and by directly providing high-resolution query tokens after projection to InstructBLIP (highResProj) for training  (Table \ref{tab:combined_fusion_analysis}). Our results show that AFR consistently outperforms these fusion strategies, highlighting its robustness and effectiveness in multimodal feature integration. Furthermore, \( AFR_{High\_res} \) offers superior computational efficiency compared to highResProj due to the lower token count (514 vs. 257 respectively) in LLM owing to feature fusion. Due to restricted space, we offer fine-grained analysis of Tab \ref{tab:combined_fusion_analysis} and results for comprehensive training on AITW for the AFR method in the appendix.

%% file: tables/computational_analysis.tex

\begin{table}
    \centering
    \resizebox{0.7\columnwidth}{!}{
        \begin{tabular}{lcc}
            \toprule
            \textbf{Method} & \textbf{TFLOPs} & \textbf{Latency}  \\
            \midrule
            MobileVLM\cite{wu2024mobilevlm} & 8.82 & 2.16 s\\
            CogAgent\cite{hong2024cogagent} &  11.86 & 3.42 s\\
            InstructBlip* & 3.19 & 0.63 s \\
            AFRAgent\textsubscript{Low-res} & 3.2 & 0.78 s\\
            AFRAgent\textsubscript{High-res} & 5.47 & 1.24 s\\
            \bottomrule
         \end{tabular}
    }
    \caption{\OURNAME achieves notably lower FLOPs and operates with substantially faster inference times compared to previous methods. * represent 257 tokens in Qformer}
    \label{tab:complexity_analysis}
\end{table}

%% file: plot_tex/grad_cam.tex

\begin{figure}[t]
  \centering
  \includegraphics[width=\linewidth]{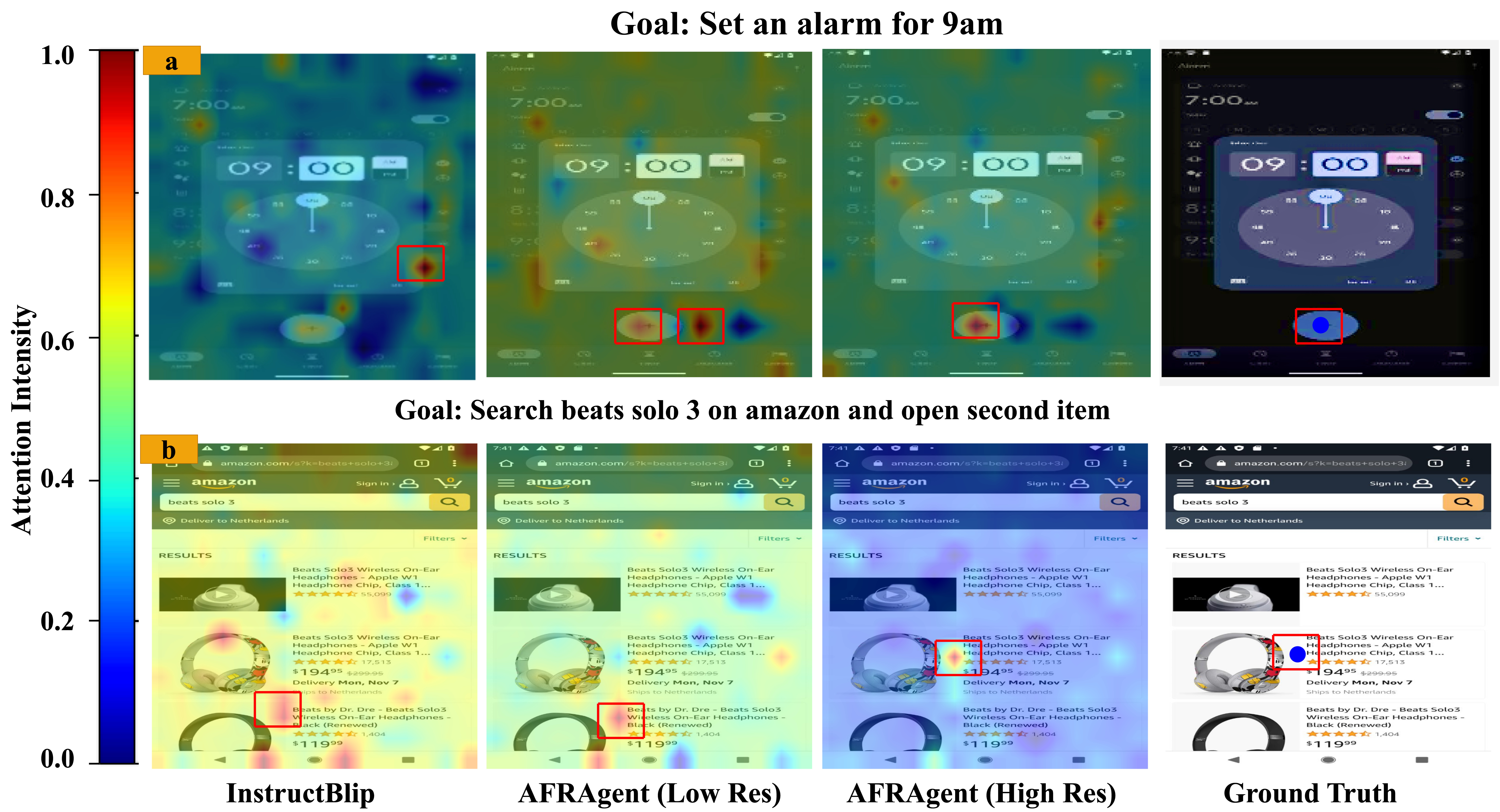}
  \caption{Grad-CAM visualizations showing AFRAgent's improved attention to task-relevant UI elements.}
  \label{fig:gradCam}
\end{figure}

%% file: tables/different_fusion.tex
\begin{table}[h!]
    \centering
    \resizebox{\columnwidth}{!}{
        \begin{tabular}{l|ccc|ccc}
            \toprule
            \textbf{Dataset} & \textbf{Residual} & \textbf{MoE} & \textbf{AFR}$_{\text{Low\_res}}$ & \textbf{highResProj} & \textbf{Qwen2-VL} & \textbf{AFR}$_{\text{High\_res}}$ \\
            \midrule
            FLOPs (T) & \textbf{3.2} & 3.54 & \textbf{3.2} & 6.46 & 17.08 & \textbf{5.47} \\
            Size (Billion) & 4.03 & 4.03 & 4.03 & \textbf{4.03} & 8.29 & \textbf{4.03} \\
            \midrule
            General & 69.61 & 69.75 & \textbf{70.2} & 69.74 & 70.15 & \textbf{70.91} \\
            Single  & 85.13  & 85.25 & \textbf{85.37} & 86.17 & 85.21 & \textbf{86.3} \\
            \bottomrule
        \end{tabular}
    }
    \caption{Comparison of AFR with prior fusion approaches in both low-res and high-res settings.}
    \label{tab:combined_fusion_analysis}
\end{table}

%% file: sec/7_conclusion.tex
\section{Conclusion}

We introduce \OURNAME, a lightweight, high-performance multimodal large language model that outperforms existing methods across multiple benchmarks. Our approach employs a novel Adaptive Feature Renormalization (AFR) technique to enrich global and high-resolution image features efficiently. We provide a comprehensive evaluation of \OURNAME through quantitative and qualitative analyses, demonstrating its effectiveness in UI automation. 